\pdfoutput=1

\documentclass[11pt]{article}

\usepackage{acl}
\usepackage{amsmath}

\usepackage{annotates}

\usepackage{comment}
\usepackage{times}
\usepackage{latexsym}
\usepackage{graphicx}
\usepackage[T1]{fontenc}
\usepackage{booktabs}
\usepackage{paralist}
\usepackage[utf8]{inputenc}

\usepackage{microtype}
\usepackage{multirow}

\newcommand{\code}[1]{\texttt{#1}}
\newcommand{\scode}[1]{\texttt{\small #1}}

\title{Meta Learning for Code Summarization}

\author{Moiz Rauf \\  Department of Computer Science \\ University of Stuttgart \\ \texttt{\small{moiz.rauf@iste.uni-stuttgart.de}}  \\
	 \And \And
  Sebastian Padó \\  Institut für Maschinelle Sprachverarbeitung \\   University of Stuttgart \\ \texttt{\small{pado@ims.uni-stuttgart.de}} \\ 
  \And \And
	Michael Pradel \\	Department of Computer Science \\	University of Stuttgart \\	\texttt{\small{michael@binaervarianz.de}}}

\begin{document}
\maketitle
\begin{abstract}
  Source code summarization is the task of generating a high-level
  natural language description for a segment of programming language
  code. Current neural models for the task differ in their
  architecture and the aspects of code they consider. In this paper,
  we show that three state-of-the-art models for code summarization work well on
  largely disjoint subsets of a large code base. This complementarity
  motivates model combination: We propose three meta-models that
  select the best candidate summary for a given code segment. The two
  neural models improve significantly over the performance of the best
  individual model, obtaining an improvement of 2.1 BLEU points on a
  dataset of code segments where at least one of the individual models
  obtains a non-zero BLEU.
\end{abstract}

\section{Introduction}

Source code summarization is the task of generating short natural
language statements describing a segment of code \cite{1,2}. Such
summaries serve an integral role in software development by aiding
code comprehension \cite{4,3}.
The recent availability of large code bases and advances in machine
learning have given this task significant attention at the interface
between NLP and software engineering. Most neural network-based
approaches build on machine translation (MT) strategies, framing code
summarization as a text-to-text generation task \cite{kuhn2017}.

A first interesting parallel to MT research in NLP is that code
summarization models also differ substantially in their assumptions
about the nature of the task. Some adopt a sequence-to-sequence
mapping approach \cite{5,Eberhart2019Automatically}, while others take
into account code structure, e.g., abstract syntax trees (ASTs)
\cite{sbt,wan2018improving,leclair2019}, or infer latent structure
with graph neural networks \cite{LeClair2019GNN} or transformers
\cite{ahmad2020summarization}. Another active direction, again
similar to many NLP tasks, is the inclusion of contextual and
background information, through API calls \cite{ijcai2018-314},
information from other methods or projects
\cite{Haque2020Filecontext,BansalHM21}, or exploiting the symmetry
between code summarization and generation \cite{wei2019code}.

\begin{figure}[tb]
	\includegraphics[width=0.5\textwidth]{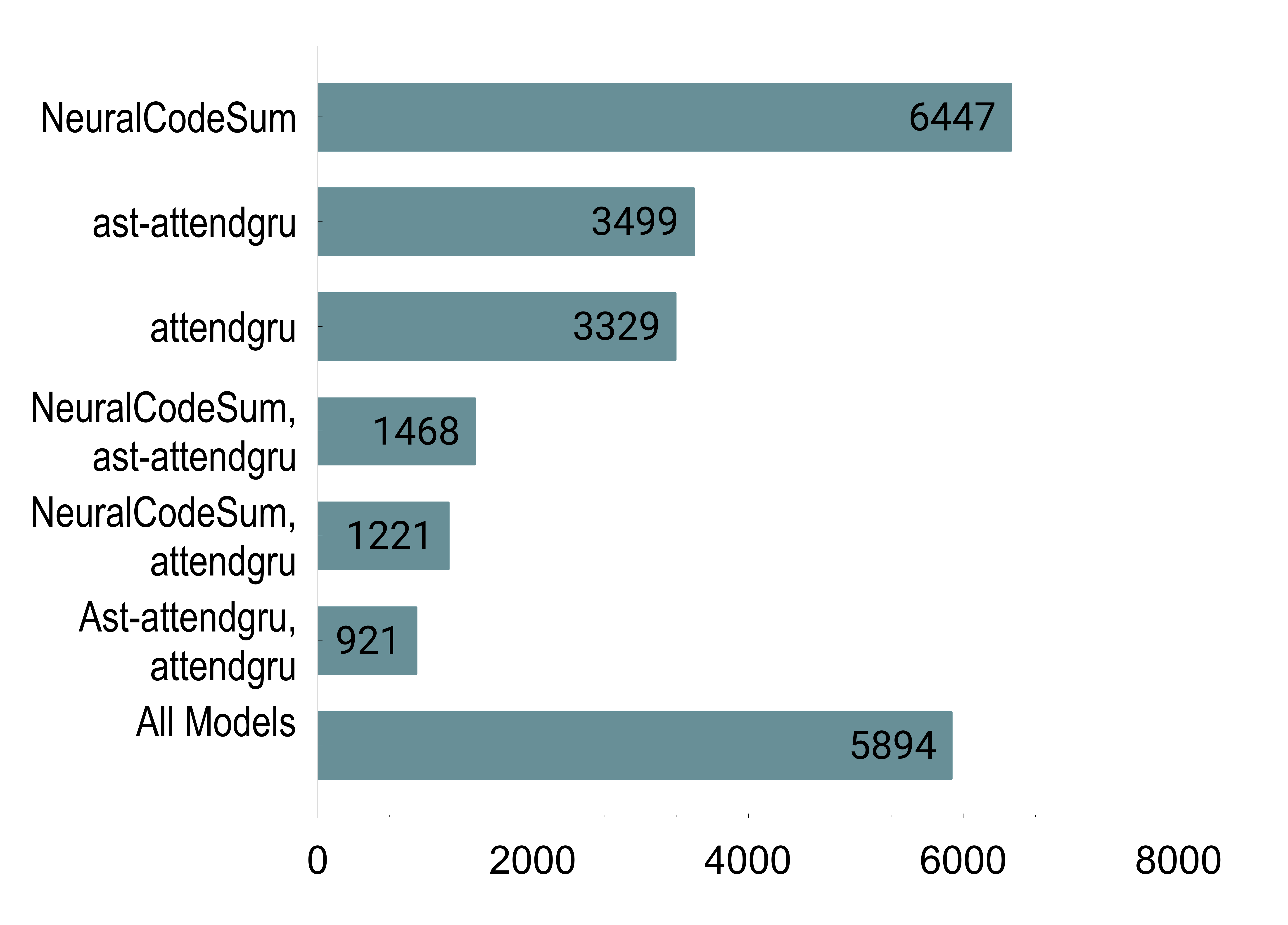}
	\caption{Complementarity of code summarization models: \#
          of FunCom methods for which each model achieves highest BLEU
          score (, indicates draw).}
	\label{fig:bldist}
\end{figure}

In this paper, we follow up on the observation by \citet{leclair2019}
that current models perform well for some examples. An analysis on three state-of-the-art methods (\emph{NeuralCodeSum},
\emph{ast-attendgru}, \emph{attendgru}, cf. Sec.~\ref{sec:smodels}) on
the Funcom dataset (Sec.~\ref{sec:data}) shows that the models are
indeed largely \textit{complementary}
(cf. Figure~\ref{fig:bldist}): Each of the individual models creates the best summary for a
substantial number of code segments, with the best model
\emph{NeuralCodeSum}, winning in about 6.4k of 22k cases where any
model predicts a summary with non-zero BLEU. Table \ref{tab:example}
illustrates this complementarity on two short methods: even though all models
learn cues from code identifiers (here, method and variable names), in
most cases they are only partially successful, and no single model is
always best.

\begin{table*}[tb]
	\centering
	\resizebox{\textwidth}{!}{
		\begin{tabular}{@{}lllr@{}}
                  \toprule
			\textbf{Code} &\textbf{Source}& \textbf{Summary}&\textbf{BLEU}\\
			\midrule
			&Reference&gets the value of the helpful votes property &\\
			\scode{public BigInteger getHelpfulVotes()\{}&\textit{NeuralCodeSum}&gets the value of the helpful votes property&1.00\\
			\scode{  return helpfulVotes;}&\textit{attendgru}& gets the value of the reason votes property&0.59\\
		\scode{\}}&\textit{ast-attendgru}&gets the value of the reason type property&0.54\\
		\midrule
						&Reference&determines whether to display the last button in the bottom pane&\\
			\scode{public void displayLastButton(boolean b) \{} &\textit{NeuralCodeSum}&display the last button&0.17\\
			\scode{ bottomPane.lastButton.setVisible(b);}	&\textit{attendgru}& displays the last button&0.00\\
				\scode{\}}&\textit{ast-attendgru}& display the last button in the panel&0.46\\
			\bottomrule
		\end{tabular}
	}
        \caption{Summaries predicted by three state-of-the-art code
          summarization models and BLEU score compared to a
          human-written reference.}
		\label{tab:example}
\end{table*}

Based on these observations, we propose to combine the strengths of
the individual code summarization models with meta learning
\cite{meta}, training a new model that selects the best summary, given
the original code segment and candidate summaries. We find a
statistically significant improvement over the best individual models.

\section{Methods}
\label{sec:methods}

Given a sequence $T=(t_1, ..., t_l)$ of code tokens, code
summarization is the task to produce a sequence $S=(w_1, ..., w_k)$ of
words describing the code. The predictions are evaluated against
reference summaries, using BLEU score~\cite{bleu} as also customary in
MT.


\subsection{Code Summarization Models}
\label{sec:smodels}
As sketched above, a number of code summarization models have been
proposed in the literature. We consider three models. All use
an encoder-decoder structure, and yield state-of-the-art results.
\begin{description}
\item[Text-based] The \emph{attendgru} model uses
  an LSTM as encoder to summarize the token sequence into a context
  vector~\cite{leclair2019}. The decoder then uses this vector to generate the summary.

\item[Code structure-based] The \emph{ast-attendgru}
  model is an extension of \emph{attendgru}~\cite{leclair2019}. In
  addition to the tokens, it also considers a flattened abstract
  syntax tree (AST). It encodes both inputs separately and feeds their
  concatenation into a decoder.

\item[Transformer-based.] The \emph{NeuralCodeSum}
  model~\cite{ahmad2020summarization} uses a transformer with relative
  positional encoding and copy attention as encoder, and then predicts
  a summary with a decoder.
\end{description}

\subsection{Meta-Learning Model}

Given the complementarity of these models
(cf. Figure~\ref{fig:bldist}), it would be very desirable to combine
their strengths. There are multiple ways to do so. Straightforward
combination of model output, as usual in ensembling \cite{Rokach2010},
is difficult for highly structured output such as
summaries. \citet{LeClair2021ENSEMBLE} combine multiple source
encoders with a joint decoder, which is effective but requires
disassembling models. In this paper, we instead adopt a meta-learning
approach \cite{meta} in which we learn a \textit{summary selector}. We
formulate this task as \emph{multi-label binary classification tasks},
where the meta-model predicts the suitability of each candidate
summary, given the summary and the original code segment. We propose
three such classifiers.


\subsubsection{Feature-Based Meta-Model}

Our first classifier, \textit{meta}$_{\text{feat}}$, is a logistic
regression model \cite{logisticregression} whose features are designed
to capture properties of code segments which may determine the
difficulty of generating code summaries, building on ideas from
performance prediction \cite{sean} and confidence estimation for
summarization \cite{louis-nenkova-2009-performance}. We consider the
following feature types:
\begin{description}
\item[Token and word frequencies]  Based on the frequency of
  each code token and each word across the codebase, we consider the
  harmonic means $\overline{\mathit{freq}}(T)$ and
  $\overline{\mathit{freq}}(S)$ of the code and summary, respectively.
  The hypothesis is that higher frequencies should make for simpler
  summarization.
		
\item[Length] We consider the number $|T|$ of tokens in a code
  segment and the number $|S|$ of words in a summary, with longer code
  segments and summaries indicating higher complexity and thus
  difficulty.

\item[Distinctiveness] We measure how distinctive a candidate summary
  $I$ is compared to all summaries produced by the same model as the
  Kullback–Leibler divergence \(\mathit{Dis}_{kl}(P_{i}||P)\), where
  \(P\) is the unigram distribution of all summaries, and
  \(P_{i}\) is the unigram distribution of candidate summary
  $i$. We expect low distinctiveness to lead to difficult
  summarization.
\end{description}

\subsubsection{Neural Meta-Models}

\begin{figure}[tb]
 	\includegraphics[width=\linewidth]{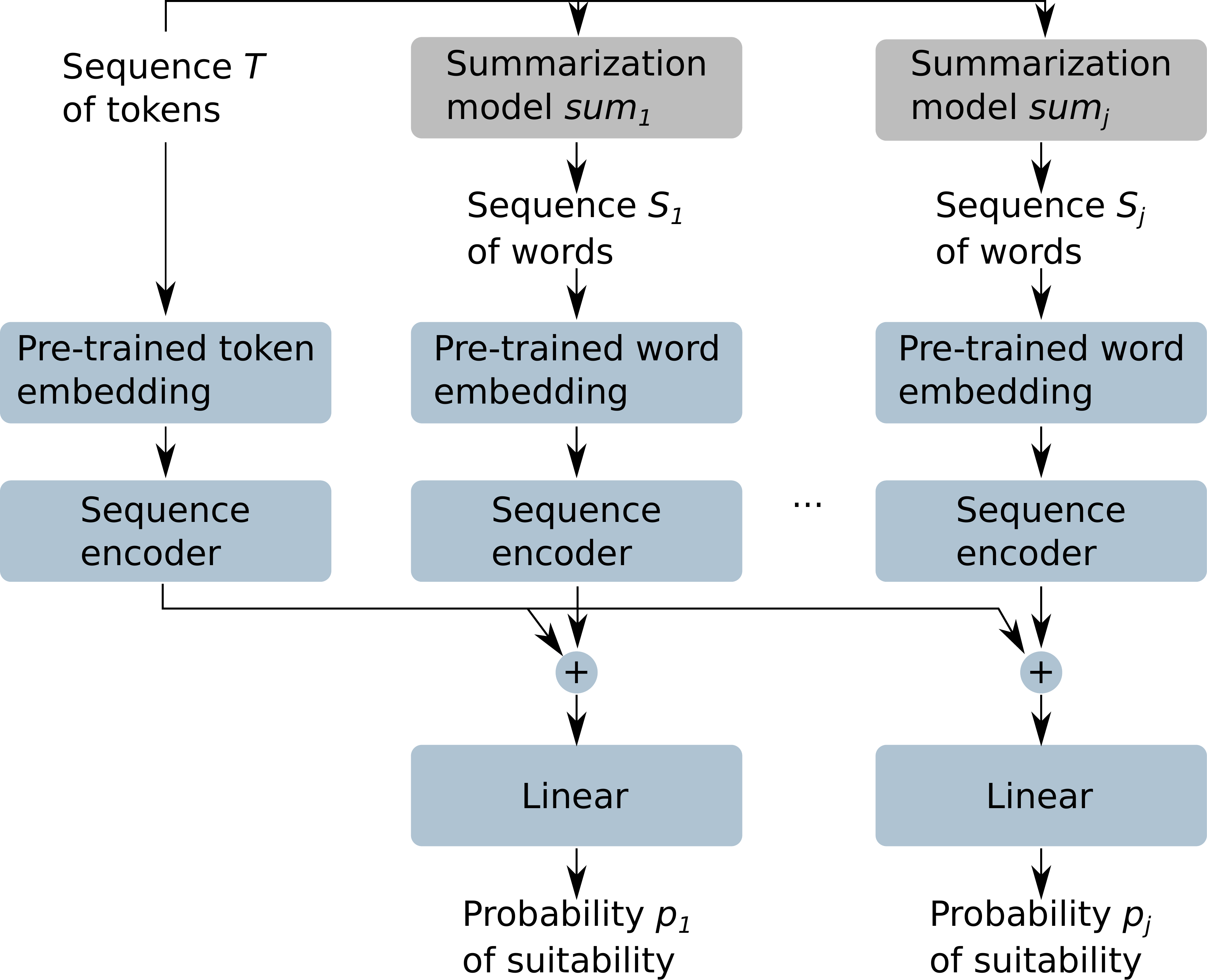}
 	\caption{Architecture of our neural meta-models.}
 	\label{fig:arch}
\end{figure}

As an alternative to specifying the relevant features by hand, we
define two neural meta-models that select a summary based on
self-learned distributed features. More specifically, as shown in
Figure~\ref{fig:arch}, we first represent the input sequences (code
tokens and summaries) in terms of FastText token and word embeddings,
respectively, The choice of FastText is motivated by prior work
showing that FastText outperforms other pre-trained token embedding
models at accurately representing identifiers in source
code~\cite{icse2021}. We pretrain these embeddings on the training
dataset used in the evaluation (see below). After embedding, the model
consists of two encoders, one for the code token sequence (generating
a vector $v_T$) and one for each summary (generating a vector $v_S$).

The final step is to concatenate, for each summary, $v_T$ and $v_S$.
The concatenation is passed through two linear layers, and finally through a sigmoid function so that each
summary is associated with a probability. The two sequence encoders
and the linear layers are trained jointly.

Our two neural models differ only in the type of sequence encoder they
use. The first model, called \textit{meta}$_{\text{LSTM}}$, encodes
sequences through a bi-directional LSTM.  The other model, called
\textit{meta}$_{\text{TRN}}$, is based on a transformer.

\subsubsection{Training and Querying Meta Models} 

As the goal of the meta-model is to maximize the overall BLEU score of the predicted summaries w.r.t.\ reference summaries, we train the meta-models in a supervised manner based on labels derived from BLEU scores.
We label a summary as suitable if and only if it achieves the best BLEU score among all available candidate summaries.
If multiple candidate summaries achieve the same, non-zero BLEU score, then all these candidates are labeled as suitable.
Let $B$ be the set of BLEU scores obtained by candidate summaries $S_1,...,S_j$ for a code sequence $T$, then the training label for $T, (S_1,...,S_j)$ is $p_1,...,p_j$ where
\[
    p_i = 
\begin{cases}
    1,& \text{if } \mathit{BLEU}(S_i, S_{\mathit{ref}}) = \mathit{max}(B)\\
    0,& \text{otherwise}
\end{cases}
\]
At inference time, we choose the candidate summary $S_i$ with the
highest predicted probability $p_i$.

\section{Experimental Setup}
\label{sec:data}

\paragraph{Data.}
We use the \emph{FunCom} dataset~\cite{dataset}. It
contains 2.1 million pairs of Java code segments and summaries, with an
average of 51 tokens per segment and 15 words per summary. We use the
authors' tokenization. As shown in Table \ref{tab:data}, we divide the
dataset into three partitions: for summary generation, for
meta-learning, and for testing. The test partition
corresponds to the one used in previous work \cite{leclair2019,LeClair2019GNN,Haque2020Filecontext,LeClair2021ENSEMBLE}, whereas the partition to train summarization models is smaller than in prior work, as we keep some data for the meta-model.
Because for a substantial percentage of code segments, all summarization models fail
to produce a summary with non-zero BLEU, we also consider a \emph{filtered} dataset
containing only segments where at least one summarization model achieves BLEU $>$ 0.
The filtered dataset hence are the cases where the meta-model has a chance to improve over
the individual models. 

\paragraph{Models and Evaluation.}
We first train the three code summarization models and then our
meta-models, as defined in Section~\ref{sec:methods}.  We
evaluate the summaries by the standard choice of corpus-level
aggregated BLEU scores \cite{bleu}. We consider three scenarios, which differ on whether the meta-model is
trained on the entire meta partition or only the filtered portion, and
analogously whether it is evaluated on the full or the filtered
portion of the test partition.

We make our code and data available. More information, along with
hyperparameters, can be found at the repository \footnote{https://github.com/sola-st/MetaCodeSum}.

\begin{table}[tb]
     \centering
   	\begin{tabular}{@{}llll@{}}
   		\toprule
   		\textbf{Partition}&\textbf{Split} &\textbf{All}&\textbf{Filtered}\\
   		\midrule
 
   		\multirow{2}{*}{Summarization}&train& 1.4 million& NA\\
   		& valid& 60k& NA \\\midrule
   		\multirow{2}{*}{Meta}& train& 440k& 101k\\
   		& valid& 70k& 5.3k\\\midrule
   	Test	& test& 101k& 22k\\
   	\bottomrule
   	\end{tabular}
   \caption{Statistics of the experiment datasets}
   \label{tab:data}
\end{table}


\section{Results}

\begin{table}[tb]
  \centering
  \renewcommand{\tabcolsep}{5pt}

	\begin{tabular}{@{}llccc@{}}
		\toprule

          &\textbf{Model} & \multicolumn{3}{c}{\textbf{Train/test of meta model}}\\
          \cmidrule{3-5}
          && All/ & All/ & Filtered/ \\
          && all & filtered & filtered \\
           \midrule
		
		\multirow{3}{*}{\rotatebox{90}{Summar.}}&\textit{attendgru}&16.25 &48.29 & 48.29\\
		&\textit{ast-attendgru}&16.62 &49.35 & 49.35\\
		&\textit{NeuralCodeSum}&18.57 &55.66 & 55.66 \\\midrule
		\multirow{3}{*}{\rotatebox{90}{Meta}}&\textit{meta}$_{\text{feat}}$& 17.93 & 52.47 & 55.06\\
		&\textit{meta}$_{\text{LSTM}}$&18.94*&57.22* & \textbf{57.08}*\\
		&\textit{meta}$_{\text{TRN}}$&\textbf{19.18}*&\textbf{57.74}* & 56.94*\\
         
		\bottomrule
	\end{tabular}
	\caption{BLEU scores on test set for individual summarization models
          and meta models. * indicates a statistically significant improvement over $NeuralCodeSum$ at $\alpha$=0.05.}
	\label{tab:eval}
\end{table}

Table \ref{tab:eval} shows our main results.  We first consider the
setup with training and test on the full dataset. Among the individual
summarization models, the transformer-based \textit{NeuralCodeSum}
model works best, with a BLEU of 18.6. Both neural meta models improve
over the individual models; the difference to \textit{NeuralCodeSum}
is statistically significant at $\alpha$=0.05. The transformer-based
meta model achieves the best result at 19.2 BLEU (+0.6 BLEU). In
contrast, the feature-based meta model even underperforms the best
individual code summary model. This highlights the difficulty of
predicting the quality of summaries for code segments, while the
quality of summaries for natural language texts has been predicted
successfully~\cite{louis-nenkova-2009-performance}.

If we evaluate the same models only on the filtered datasets -- i.e.,
where the meta model has a chance of improving over the individual
models (middle column) -- we observe the same ranking of the models,
but the margin between the best individual summarization model
(\textit{NeuralCodeSum}, 55.7 BLEU) and the neural meta learning
models has increased: We obtain a BLEU of 57.2 for the $\mathit{meta}_{\mathit{LSTM}}$
(+1.5 BLEU) and a BLEU of 57.7 for the $\mathit{meta}_{\mathit{TRN}}$ (+2 BLEU);
differences again are significant at $\alpha=0.05$. We take these numbers
as an indication that the neural meta-learning approach is
generally successful for code segments for which ``sensible''
summary candidates (with BLEU $>$ 0) have been produced by the
individual models.

Finally, the right-hand column assesses the consequences of training
the meta-models only on such ``sensible'' summary candidates exist.
Compared to the middle column, the meta-model results are slightly
lower.
In other words, the apparently
uninformative summary candidates still contribute to the success of
the meta model. Taken together with the observation that the results
for the BiLSTM changes much less (-0.1 BLEU) than for the transformer
(-0.8 BLEU), we propose the following interpretation: Pairs of code
segments and non-sensical summaries may still help the neural model in
learning to encode typical code token and word sequences, which is
more important for the transformer, with its higher capacity, than for
the BiLSTM. 

\section{Conclusions}

The present paper exploits the complementary nature of different code
summarization models through a meta-learning approach.  We find that
neural models can predict the best summary from a set of candidates
created by three state-of-the-art models, yielding an increase in BLEU
of up to 2.1 points. We believe our results to be promising, and
future improvements of individual summarization models will give our
meta-models better predictions to choose from. At the same time, our
results also highlight directions for future work, including
meta-model introspection (why does the transformer succeed where the
manual features fail?) and a re-evaluation of BLEU as summary evaluation
metric~\cite{10.1162/tacl_a_00373}.

\bibliography{bibliography}
\bibliographystyle{aclnatbib}

\appendix
\end{document}